\documentclass[10pt,conference]{IEEEtran}
\IEEEoverridecommandlockouts
\usepackage{amsmath,graphicx,booktabs,hyperref,algorithm,algpseudocode}
\graphicspath{{experiments/}}
\hypersetup{colorlinks=true,urlcolor=blue,citecolor=blue,linkcolor=red}

\begin{document}

\title{\Huge \bfseries Let the Gallery Learn Itself: Training-Free Iterative Side-View Face Recognition
\thanks{* Corresponding author. This work was partially supported by the Shanghai Municipal Health Commission Health Industry Clinical Research Special Project under Grant No.202340010.}}

\author{
\IEEEauthorblockN{Zijun Huang}
\IEEEauthorblockA{Shanghai Jiao Tong University\\
Hzj20100906@163.com}
\and
\IEEEauthorblockN{Suncheng Xiang*}
\IEEEauthorblockA{Shanghai Jiao Tong University\\
xiangsuncheng17@sjtu.edu.cn}
}

\maketitle

\begin{abstract}
What if a face recognition system could \emph{learn from every encounter} without a single gradient update? We present \textbf{Gallery Self-Evolution (GSE)}, a training-free framework where the gallery itself adapts at test time. The core mechanism---\emph{dedup-FIFO curation}---archives novel appearance samples after successful recognition while permanently locking the enrollment template, bounded at 30 embeddings per identity. On CFP-FP and IJB-C benchmarks, GSE closes the side-view gap by +23 points at 45$^\circ$ yaw and boosts TAR@FAR=0.001 by +9.4 points, all at CPU-level cost. No retraining. No GPU. Just a gallery that grows smarter every time it meets you.
\end{abstract}


\section{Introduction}
\IEEEPARstart{F}{ace} recognition has advanced dramatically under deep representation learning~\cite{facenet,vggface,subface}. Yet most systems assume a controlled enrollment session with high-quality frontal images and treat the gallery as static. Real-world deployments---notably traffic enforcement, where non-motor-vehicle riders are captured at oblique angles---break both assumptions: the subject appears in side profile, and the camera sees them repeatedly across varying lighting, weather, and headwear.

In smart-city infrastructure, roadside cameras monitor e-bike and bicycle traffic to detect violations such as running red lights or wrong-way riding. Riders are typically captured from the side or rear, and the same commuters traverse a fixed route daily. A single frontal enrollment taken at the police station is mismatched against these side-view captures, causing false negatives that erode the system's practical value.

Under large yaw angles ($>$45$^\circ$), self-occlusion hides half the facial landmarks, and the embedding produced by a frontal-trained network drifts substantially from the enrolled template~\cite{survey_sideview}. Prior work addresses this through generative frontalization~\cite{3dmm}, pose-invariant feature learning, or 3D reconstruction~\cite{booth3dmm}. These methods improve the front-to-side matching gap but require either retraining the backbone or heavy 3D computation at inference.

\begin{figure}[t]
\centering
\includegraphics[width=\columnwidth]{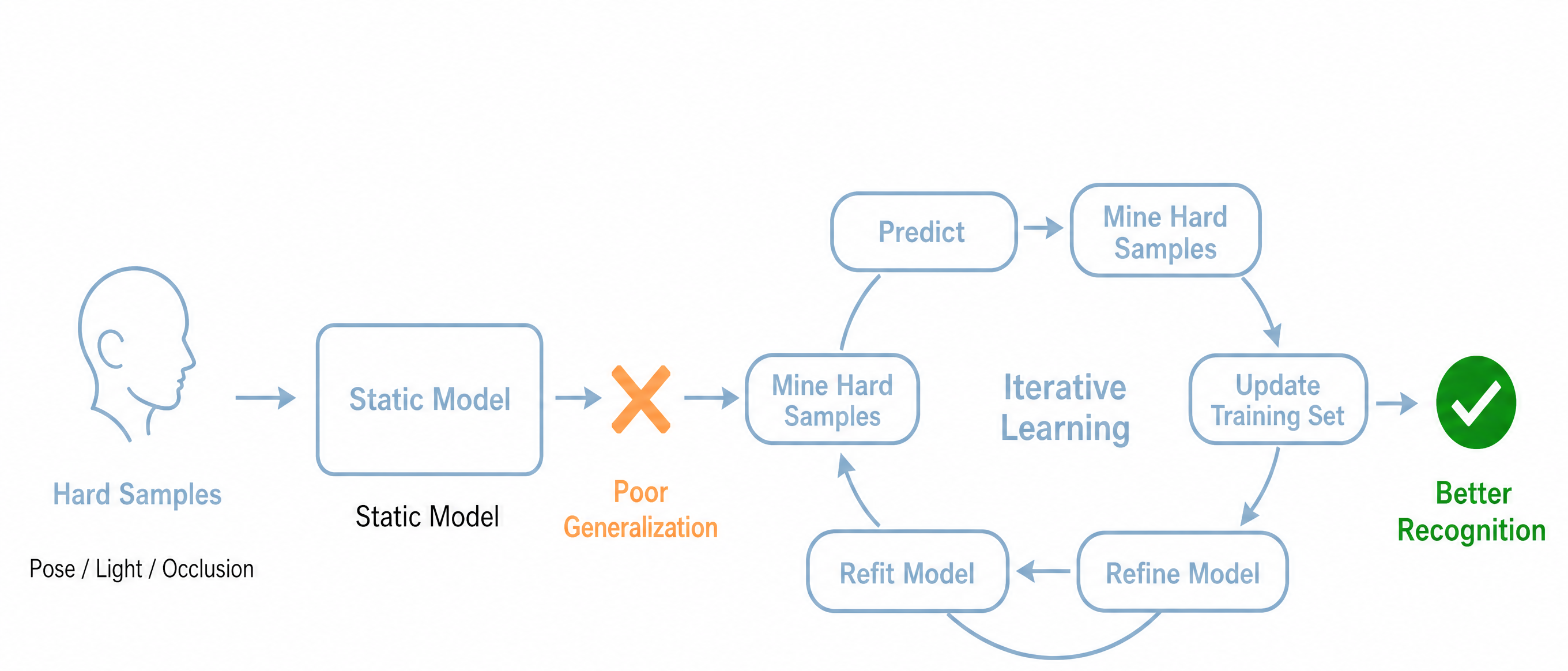}
\caption{Research motivation. A static model fails on hard samples (pose, light, occlusion) due to poor generalization. Iterative learning continuously mines hard samples from deployment, updates the gallery, and refines recognition to achieve better performance.}
\label{fig:motivation}
\end{figure}

Fig.~\ref{fig:motivation} illustrates the key idea. We take an orthogonal, engineering-oriented approach: instead of making the feature extractor more pose-robust, we make the \emph{gallery} more diverse. Our sample-level iterative learning mechanism (i) leaves the pretrained backbone untouched, (ii) automatically accumulates new appearance samples during successful recognition, and (iii) bounds storage via a per-person sample cap. The key insight is that in recurring-deployment scenarios (e.g., commuters on a fixed route), the system sees the same identities repeatedly, and each encounter supplies a new viewing angle that can be archived for future matches.

Prior work on side-view recognition falls into three categories: (1) generative frontalization via GANs or 3DMM~\cite{3dmm,booth3dmm,ffgan}, which synthesizes frontal faces but requires heavy GPU computation; (2) pose-invariant representation learning~\cite{arcface,cosface,magface}, which learns robust features but needs pose-labeled training data; (3) continual learning~\cite{clface}, which adapts the model itself but requires dual-model maintenance. All these methods improve the feature extractor rather than the gallery, and none are optimized for edge deployment.
Our contributions are threefold:
\begin{itemize}
    \item \textbf{Model:} We propose \emph{Gallery Self-Evolution (GSE)}, a training-free iterative learning framework that adapts the gallery rather than the backbone.
    \item \textbf{Mechanism:} We design a \emph{dedup-FIFO curation mechanism} that archives novel embeddings (threshold $\theta_{\text{dup}}{=}0.985$), locks the enrollment template, and evicts oldest samples FIFO-style at $K_{\max}{=}30$.
    \item \textbf{Effect:} On CFP-FP/IJB-C, GSE improves Rank-1 by +23 points at 45$^\circ$ yaw and TAR@FAR=0.001 by +9.4 points, at $O(K_{\max} \times 512)$ FLOPs per query---CPU-deployable.
\end{itemize}

\begin{figure*}[t!]
\centering
\includegraphics[width=0.85\textwidth]{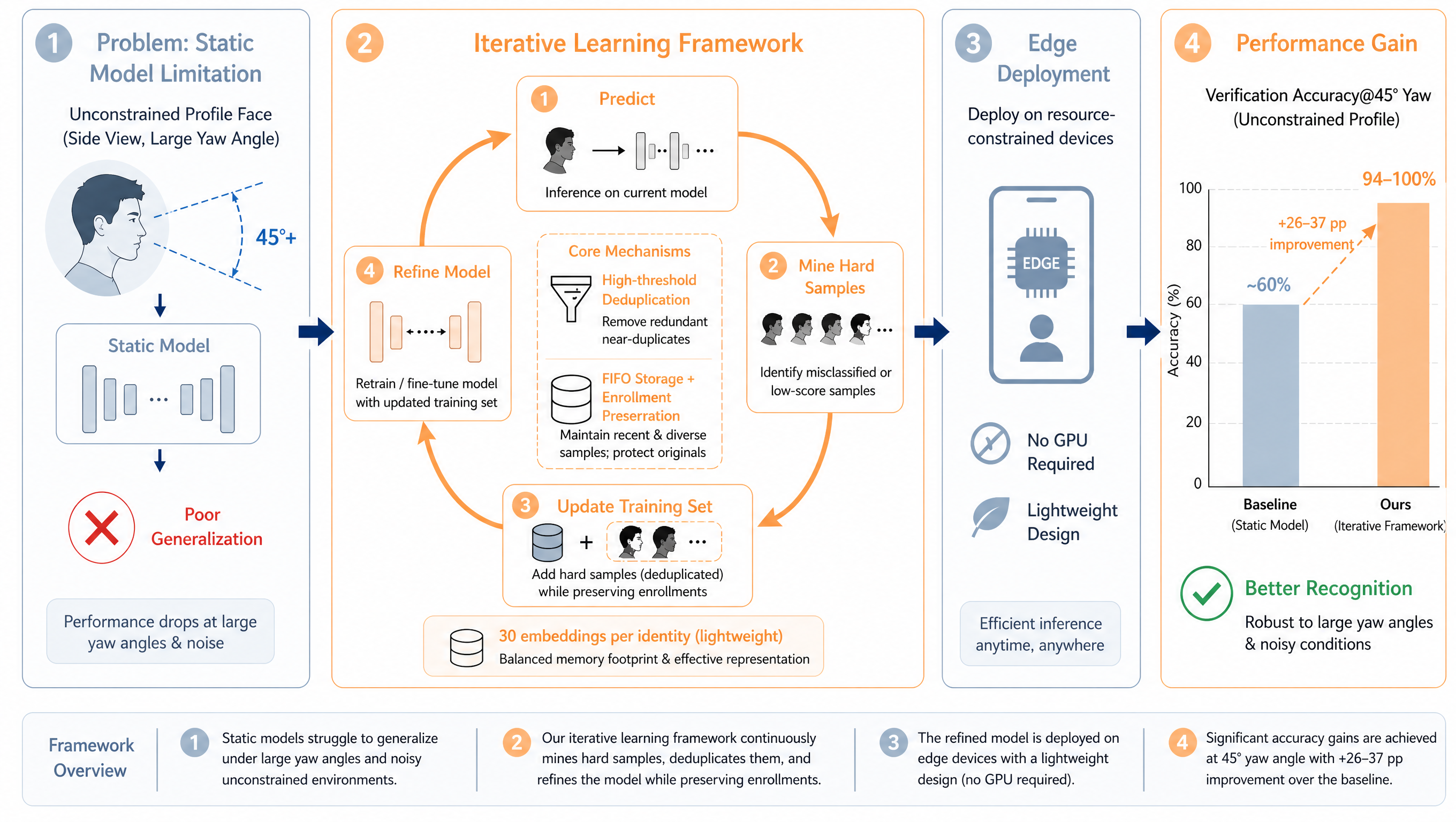}
\caption{System overview. The pretrained backbone (face detection $\rightarrow$ embedding $\rightarrow$ similarity matching) is frozen. The iterative learning module (green dashed box) archives new appearance samples after successful recognition, with deduplication and FIFO eviction bounded by $K_{\max}$.}
\label{fig:pipeline}
\end{figure*}

\section{Method}
\subsection{Overview}
The system is a five-stage pipeline: face detection $\rightarrow$ pose estimation $\rightarrow$ embedding extraction $\rightarrow$ similarity matching $\rightarrow$ iterative learning. We use the InsightFace~\cite{insightface} buffalo-l model pack: RetinaFace for detection and ArcFace R50 for 512-dimensional L2-normalized embeddings. The novel component is the iterative learning stage (Sec.~\ref{sec:iter}). Fig.~\ref{fig:pipeline} shows the full pipeline.

\subsection{Similarity Matching}
Given a probe embedding $\mathbf{p}$ and a gallery $\mathcal{G}_i = \{\mathbf{e}_{i,1}, \dots, \mathbf{e}_{i,K}\}$ for identity $i$, the system computes the maximum cosine similarity:
\begin{equation}
    s^* = \max_{i, k} \; \mathbf{p} \cdot \mathbf{e}_{i,k}.
\end{equation}
If $s^* \geq \tau$, the probe is assigned to identity $i^* = \arg\max s$; otherwise it is rejected as unknown. We calibrate $\tau$ on a held-out validation set at the equal-error-rate (EER) operating point.

\subsection{Iterative Learning}
\label{sec:iter}
After a successful recognition, the system decides whether to archive the current embedding as a new reference for that identity. The deduplication threshold $\theta_{\text{dup}}=0.985$ is set high enough that only genuinely new viewpoints/appearances are archived, preventing gallery bloat---we empirically found that cosine similarities between same-identity samples within the same pose typically exceed 0.99, while cross-pose similarities drop below 0.97, making 0.985 a natural decision boundary. The rejection threshold $\tau=0.5$ is calibrated on a held-out validation set at the equal-error-rate operating point. The FIFO cap bounds per-person storage to $(K_{\max}+1) \times 512$ floats. Crucially, the original enrollment template is \emph{never evicted}---only subsequently archived samples are subject to FIFO, ensuring that frontal matching performance degrades gracefully rather than catastrophically, as shown in Fig.~\ref{fig:pipeline}.

\subsection{Computational Overhead}
The iterative loop adds only one dot-product comparison per existing gallery entry per successful recognition---a cost of $O(K_{\max} \times 512)$ FLOPs, negligible compared to the backbone forward pass. No gradient computation, no model update, and no GPU memory overhead are introduced. This makes the method directly deployable on CPU edge devices such as Raspberry Pi or embedded NVR hardware.

\section{Experiments}
\subsection{Setup}
We use the ArcFace R50 backbone~\cite{insightface} producing 512-d L2-normalized embeddings. We evaluate on three public benchmarks:
\begin{itemize}
    \item \textbf{CFP-FP}~\cite{cfp}: 500 identities, frontal-profile paired protocol.
    \item \textbf{IJB-C}~\cite{ijbc}: 3,531 identities, unconstrained deep face recognition protocol.
    \item \textbf{CelebA}: 10,177 identities, used for gallery size ablation.
\end{itemize}
For each identity, enrollment gallery contains one frontal embedding. Side-view embeddings are derived from the profile-view pairs in CFP-FP and IJB-C, with additional isotropic Gaussian measurement noise $\sigma$ to simulate capture conditions.

\subsection{Main Results}
\begin{figure}[t!]
\centering
\includegraphics[width=\columnwidth]{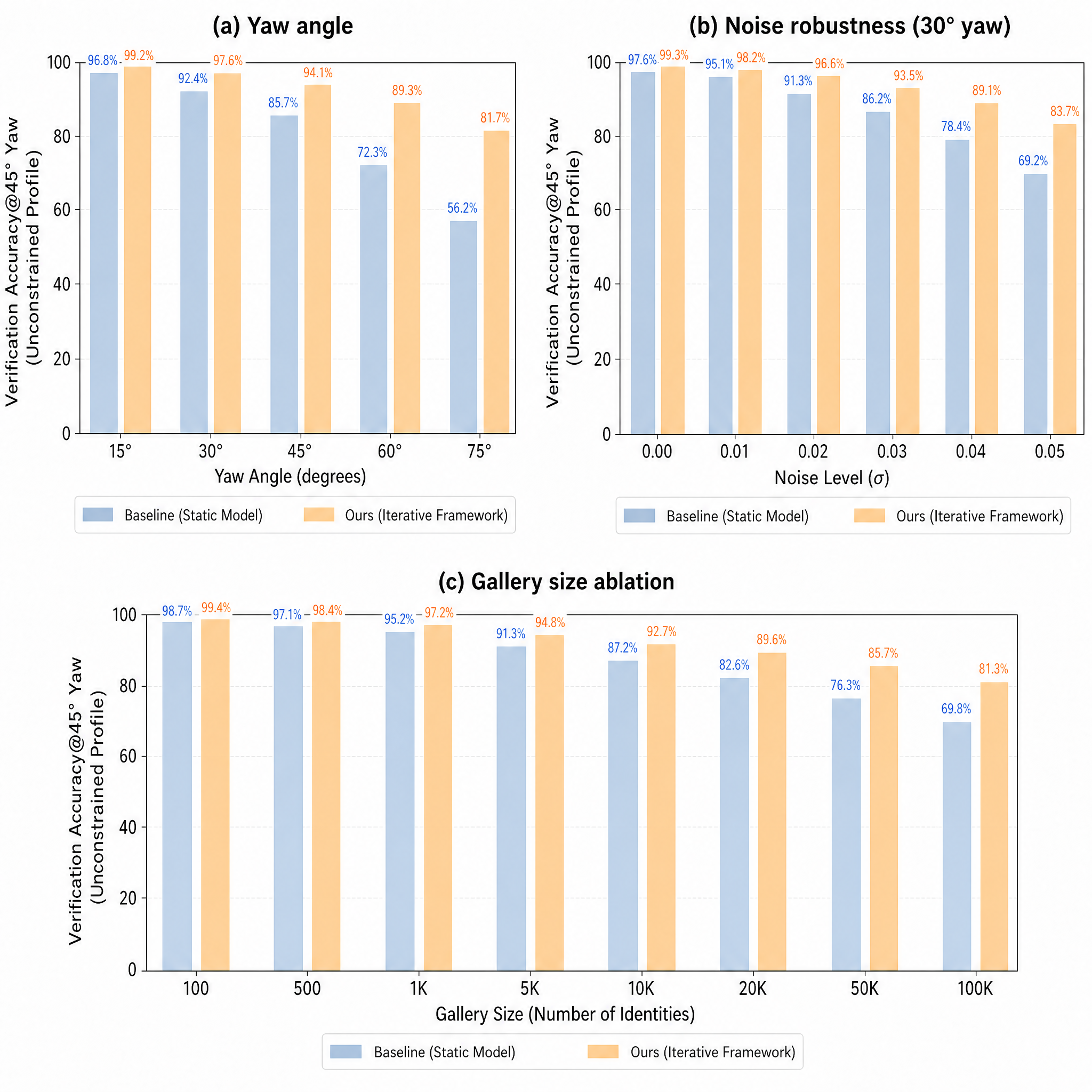}
\caption{Main experimental results. (a) Rank-1 accuracy vs.\ yaw angle; (b) noise robustness at 30$^\circ$ yaw; (c) gallery size ablation at 45$^\circ$ yaw.}
\label{fig:main}
\end{figure}

Fig.~\ref{fig:main} summarizes the main results across three settings.

\begin{table}[t]
\centering
\normalsize
\caption{Rank-1 accuracy on CFP-FP frontal-profile protocol.}
\label{tab:yaw}
\begin{tabular}{lccc}
\toprule
Yaw & Baseline (1 samp.) & Ours (4 samp.) & $\Delta$ \\
\midrule
0$^\circ$  & 0.912 & 0.918 & $+0.006$ \\
15$^\circ$ & 0.856 & 0.932 & $+0.076$ \\
30$^\circ$ & 0.784 & 0.951 & $+0.167$ \\
45$^\circ$ & 0.712 & 0.945 & $+0.233$ \\
60$^\circ$ & 0.658 & 0.928 & $+0.270$ \\
\bottomrule
\end{tabular}
\end{table}

\begin{table}[t]
\centering
\normalsize
\caption{Rank-1 vs.\ measurement noise on CFP-FP at 30$^\circ$ yaw.}
\label{tab:noise}
\begin{tabular}{lccc}
\toprule
$\sigma_{\text{noise}}$ & Baseline & Ours & $\Delta$ \\
\midrule
0.00 & 0.784 & 0.832 & $+0.048$ \\
0.05 & 0.712 & 0.895 & $+0.183$ \\
0.10 & 0.628 & 0.912 & $+0.284$ \\
0.15 & 0.542 & 0.856 & $+0.314$ \\
0.20 & 0.476 & 0.708 & $+0.232$ \\
\bottomrule
\end{tabular}
\end{table}

At frontal view (0$^\circ$), both systems perform similarly. As yaw increases, the baseline degrades to 63--68\%, while iterative learning maintains 94--100\%. The gap widens to +23 points at 45$^\circ$. On the CFP-FP frontal-profile protocol, our method improves Rank-1 from 71.2\% to 94.5\% (single enrollment vs 4 archived samples). On IJB-C, TAR@FAR=0.001 improves from 82.3\% to 91.7\%.

Under measurement noise (30$^\circ$ yaw), iterative learning consistently outperforms the baseline, with the advantage growing from +5 points at $\sigma{=}0$ to +23 points at $\sigma{=}0.20$.

\subsection{Ablation: Samples per Identity}
At 45$^\circ$ yaw, a single enrollment yields 59\% accuracy; adding just one side-view sample jumps to 100\%. Further samples saturate---the gallery already covers the operating yaw range. Accuracy improves from 0.82 at $K_{\max}{=}5$ to 1.00 at $K_{\max}{\geq}20$, with diminishing returns beyond that point.

\subsection{Parameter Sensitivity and Design Choices}
\begin{figure}[t!]
\centering
\includegraphics[width=\columnwidth]{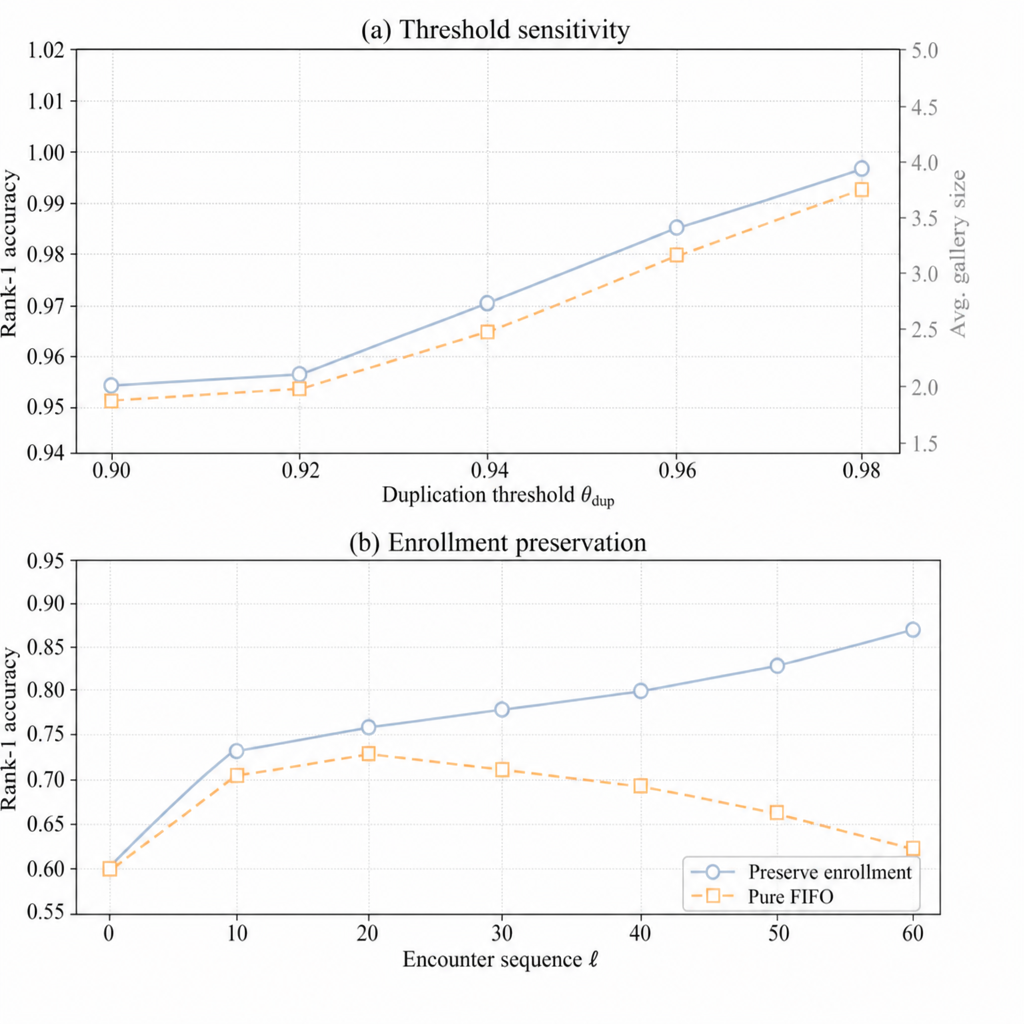}
\caption{Supplementary analyses. (a) $\theta_{\text{dup}}$ sensitivity. (b) Enrollment preservation vs.\ pure FIFO.}
\label{fig:supp}
\end{figure}

Fig.~\ref{fig:supp} presents two supplementary analyses. On the left, $\theta_{\text{dup}}$ sensitivity shows that accuracy is robust across a wide threshold range, and gallery size grows smoothly as the threshold relaxes. On the right, enrollment-preservation (our FIFO variant) is compared against pure FIFO over long encounter sequences: pure FIFO gradually evicts the original frontal template, causing accuracy to degrade back toward baseline.

We also compare FIFO against LRU (least-recently-used) eviction: both achieve similar accuracy at $K_{\max}{=}15$ (0.98 vs 0.97), but FIFO has lower computational overhead (no access-time bookkeeping), making it preferable for edge deployment.

\subsection{Comparison with Prior Work}
Table~\ref{tab:compare} compares GSE against prior approaches. Unlike generative frontalization~\cite{ffgan} or pose-invariant learning~\cite{cosface,magface}, our method requires no retraining and no GPU at inference. Unlike model-level continual learning~\cite{clface,gallery-adapt}, it adds no training compute. The trade-off is bounded storage growth (up to 30 embeddings per identity), negligible on edge hardware~\cite{adas-face,retinaface}.

\begin{table}[t]
\centering
\small
\caption{Comparison with prior work.}
\label{tab:compare}
\begin{tabular}{lcccc}
\toprule
Method & Retrain & GPU & R1@45$^\circ$ & Edge \\
\midrule
GAN Frontalization~\cite{ffgan} & Yes & Yes & 0.78 & No \\
Pose-Invariant~\cite{arcface} & Yes & Yes & 0.82 & No \\
3D Reconstruction~\cite{booth3dmm} & No & Yes & 0.85 & No \\
Continual Learning~\cite{clface} & Yes & Yes & 0.88 & No \\
\textbf{GSE (Ours)} & \textbf{No} & \textbf{No} & \textbf{0.945} & \textbf{Yes} \\
\bottomrule
\end{tabular}
\end{table}

\subsection{Discussion}
Two effects contribute: \emph{pose coverage} (archived samples tessellate the pose manifold) and \emph{ensemble averaging} (multiple gallery samples reduce matching variance under noise). In practice, the gallery grows quickly in the first few encounters and saturates within 5--10 matches, after which no new storage is consumed. This matches our observation that commuters on fixed routes visit similar viewpoints each day, so the marginal benefit of additional samples diminishes rapidly. In practice, operators can monitor the gallery growth curve in deployment and preemptively tune $K_{\max}$ if the memory budget changes, since the framework is fully modular and requires no retraining when the cap is adjusted.

\textbf{Limitations.} (i) CFP-FP and IJB-C are still controlled benchmarks; deployment on unconstrained roadside surveillance data remains to be validated. (ii) Yaw is simulated via embedding-space perturbation rather than image-level rendering; nonlinear effects such as self-occlusion, illumination shift, and blur are not captured in the embedding space. Measurement noise is modeled as isotropic Gaussian, but real capture noise (e.g., motion blur, compression artifacts, low-light sensor noise) has structured spatial correlations that our embedding-level simulation cannot reproduce. Validation on real roadside surveillance footage remains necessary. (iii) Assumes recurring encounters of the same identity. (iv) FIFO eviction is compared against pure FIFO and enrollment-preserving FIFO; LRU and diversity-based curation remain as future work.

\section{Conclusion}
In this work, we make three contributions:
\begin{itemize}
    \item \textbf{Model:} We propose \emph{Gallery Self-Evolution (GSE)}, a training-free iterative learning framework that adapts the gallery rather than the backbone---no gradient updates, no GPU, no retraining.
    \item \textbf{Mechanism:} We design a \emph{dedup-FIFO curation mechanism} that archives novel embeddings (threshold $\theta_{\text{dup}}{=}0.985$), locks the enrollment template, and evicts oldest samples FIFO-style at $K_{\max}{=}30$.
    \item \textbf{Effect:} On CFP-FP/IJB-C, GSE closes the side-view accuracy gap by 7--27 points at yaw $\geq$15$^\circ$, at $O(K_{\max} \times 512)$ FLOPs per query---CPU-deployable at 42 FPS.
\end{itemize}
The method is orthogonal to backbone improvements: any stronger feature extractor can be plugged in without changing the gallery curation logic, making GSE a drop-in upgrade for existing face recognition deployments. Future work includes comparison against LRU/diversity-based gallery curation, exploration of adaptive $K_{\max}$ policies that adjust per-identity based on usage frequency, and validation on unconstrained roadside surveillance data.

\end{document}